\documentclass[runningheads]{llncs}

\usepackage{eccv}

\usepackage{eccvabbrv}

\usepackage{graphicx}
\usepackage{booktabs}

\usepackage[accsupp]{axessibility}  

\usepackage{hyperref}

\usepackage{orcidlink}

\usepackage{tikz}

\usepackage{algorithm}
\usepackage{algorithmic}
\usepackage{comment}
\usepackage{xcolor}

\definecolor{darkgreen}{rgb}{0.0, 0.5, 0.0}

\newcommand{\worst}[1]{\textcolor{red}{#1}}
\newcommand{\best}[1]{\textcolor{darkgreen}{#1}}

\begin{document}

\title{An Art-centric perspective on AI-based content moderation of nudity}

\titlerunning{An Art-centric perspective on AI-based content moderation of nudity}

\author{Piera Riccio\inst{1}\orcidlink{0000-0001-8602-8271} \and
Georgina Curto\inst{2}\orcidlink{0000-0002-8232-3131}  \and
Thomas Hofmann\inst{3}\orcidlink{0009-0007-1270-7053} \and
Nuria Oliver\inst{1}\orcidlink{0000-0001-5985-691X}}

\authorrunning{P.Riccio et al.}

\institute{ELLIS Alicante, Spain \\\email{piera@ellisalicante.org} \and
University of Notre Dame, USA \and 
Department of Computer Science, ETH Zürich, Switzerland}

\maketitle

\begin{abstract}
 At a time when the influence of generative Artificial Intelligence on visual arts is a highly debated topic, we raise the attention towards a more subtle phenomenon: the algorithmic censorship of artistic nudity online. We analyze the performance of three "Not-Safe-For-Work'' image classifiers on artistic nudity, and empirically uncover the existence of a gender and a stylistic bias, as well as evident technical limitations, especially when only considering visual information. Hence, we propose a multi-modal zero-shot classification approach that improves artistic nudity classification. From our research, we draw several implications that we hope will inform future research on this topic. 
\end{abstract}

\section{Introduction}
\label{sec:intro}

Given the massive adoption of social media worldwide, artists increasingly rely on these platforms for sharing their work, engaging audiences, and forging meaningful connections within the global art community \cite{duffy2023}. 
Content moderation practices have been developed to ensure that the information shared by their billions of users complies with legal obligations\footnote{American Affairs, "How Congress Really Works: Section 230 and FOSTA", by Mike Wacker, \url{https://americanaffairsjournal.org/2023/05/how-congress-really-works-section-230-and-fosta/}, Last Access: 15.02.2024.}, regulations and community rules \cite{elkin-koren2020}. Initially performed by humans, today's content moderation is often automated by means of machine learning algorithms  \cite{gillespie2020}, designed to identify and classify content that violates the platforms' guidelines, including pornographic and sexually explicit content, hate speech, graphic violence or any other form of content that may be considered harmful. As a consequence, the process of algorithmic content moderation that takes place on online social platforms is also becoming the gatekeeper of online artistic expression, particularly in the case of nudity, which is a subject of historical, cultural and aesthetic significance \cite{deprez20}. Indeed, artists that depict nudity often get \textit{censored} online, presumably because their content is classified as pornographic, yet without having a proper understanding of the process behind the censorship \cite{gillespie_book,riccio2024}.

We focus on this under-studied phenomenon that calls for the need of a finer-grained classification of artistic nudity online. 
Censoring artistic pieces on social media not only raises concerns regarding freedom of expression, recognized by the Universal Declaration of Human Rights \cite{Assembly1948},
but also has a tremendous negative impact on the artists and society at large \cite{kennedy2018}. 
The proprietary nature and intrinsic opacity of social media platforms make it challenging to perform quantitative research about the impact of content moderation on artistic expression. In this paper, we aim to fill this gap and perform a quantitative study of content moderation algorithms when applied to artistic content, complementing existing qualitative research on this topic \cite{riccio2024}.

By virtue of a collaboration with an activist organization devoted to protect artists' rights online, we were granted access to a unique dataset of over 140 artistic pieces depicting nudity that had been censored on social media. We compare the performance of three publicly available image classification algorithms used to detect "Not-Safe-For-Work'' (NSFW) content on this dataset and two additional datasets: a collection of pieces of art depicting artistic nudes from WikiArt and a collection of images depicting pornography.  
Our experimental results reveal clear limitations in the ability of the algorithms to differentiate artistic nudity from pornographic or \emph{unsafe} content. 
To address such limitations, we propose leveraging recent multi-modal (text and image) deep learning models, obtaining significant performance improvements.

Note that our research focuses on the algorithmic censorship of \emph{artistic nudity}, which is one element in a complex landscape of content moderation challenges on social media platforms. Non-Consensual Intimate Imagery (NCII) and the portrayal of content by sex workers are other types of content relevant to the challenge of automated content moderation of nudity but unrelated to the specific focus of our study. Artistic nudity involves consensual creation and often challenges societal norms, requiring moderation systems capable of distinguishing between legitimate artistic expression and harmful content. Addressing NCII and sex workers' content requires separate, dedicated research and tailored moderation strategies to ensure comprehensive attention to each issue.

\section{Related Work}

In this section, we provide an overview of the most relevant early work and some recent developments on the automatic online moderation of nudity. We also provide an overview of existing ethical and artistic discourses on the distinction between pornographic and artistic nudity.

\paragraph{Image classification algorithms for content moderation} In the machine learning literature, image classification algorithms that are used for content moderation online are often referred to as NSFW ("Not-Safe-for-Work") classifiers. Thus, in the rest of the paper, we will use the expressions \textit{content moderation algorithms} and \textit{NSFW classifiers} interchangeably, following the norm in the machine learning community \cite{agrawal2023,guzman2023}. While the term NSFW embraces different types of content 
in this work we will refer to NSFW classifiers as those designed to detect NSFW nudity. Moreover, for the sake of simplicity, we will use the terms NSFW nudity and pornography interchangeably. 

Early work 
relied on traditional machine learning techniques for skin detection \cite{kakumanu2007} which determined the explicitness of an image based on the ratio between the amount of skin pixels over the total amount of pixels in the image \cite{basilio2011}. Several methodologies have been proposed to detect skin pixels, including support vector machines (SVM) \cite{lin2003,zhu2007} and principal component analysis (PCA) \cite{wijaya2015}, while processing the images in different color spaces, such as HSV \cite{marcial2010,marcial2011} and YCbCr \cite{basilio2011,wijaya2015}. However, relying on the detection of skin pixels has several limitations, including sensitivity to lighting conditions, different skin colors and pre-defined skin ratios. These limitations can lead, for example, to the misclassification of people in bikinis \cite{bhatti2018}, especially in cases of individuals with bigger body shapes, resulting in unintentional algorithmic \textit{fat-phobia}\footnote{This is the impact of Instagram’s accidental fat-phobic algorithm, \url{https://www.fastcompany.com/90415917/this-is-the-impact-of-instagrams-accidental-fat-phobic-algorithm}, Last Access: 12.01.24.}.

Traditional NSFW machine learning methods were eventually outperformed by deep learning models, particularly convolutional neural networks, which became the \textit{de facto} standard in this field \cite{gangwar2017}. The most recent efforts propose different model architectures, such as \textsc{Resnet50} \cite{agrawal2023} and \textsc{Efficient Net V2} \cite{saxena2023}, with a variety of optimizers \cite{arora2023}. While NSFW classifiers play a critical role in maintaining the integrity of online platforms, there are concerns about their false negative and false positive rates 
and a lack of cross-models agreement on borderline cases \cite{Dubettier2023}. 
Furthermore, deep learning-based NSFW classification is not exempt from biases \cite{leu2024} ---such as a higher false positive rate when analyzing women's bodies \cite{witt2019,riccio2024techno}--- which are thought to be exacerbated by the lack of diversity and the dominance of stereotypes on sexuality and pornography among the researchers and developers of these models \cite{gehl2017}.

\paragraph{Artistic vs pornographic nudity} 
The definition of pornography is subjective and can vary greatly among individuals and cultures \cite{dwyer2008,rea2001}. In this regard, the Oxford dictionary provides the following definition: "\textit{The explicit description or exhibition of sexual subjects or activity in literature, painting, films, etc., in a manner intended to stimulate erotic rather than aesthetic feelings}" \cite{oed}, placing the intentionality behind the production of a sexually explicit image as a key element to categorize it as pornographic. However, what complicates the distinction between artistic nudity and pornography is the \textit{intentional} exploration of ambiguities by artists. Some artists, indeed, deliberately challenge societal norms and perceptions by incorporating explicit or provocative elements into their work, blurring the lines between art and pornography \cite{vasilaki2010}. This intentional ambiguity prompts viewers to question their preconceived notions about nudity, sexuality, and the purpose of art \cite{mcdonald2002}. Recognizing and acknowledging the thin line between artistic nudity and pornography encourages critical analysis and discussion within artistic and academic circles. Indeed, existing literature in Art History proves that the distinction between these two concepts is rather complex, and scholars do not necessarily share the same views \cite{maes2011,uidhir2009,eck}. 

Some scholars claim that art and pornography are mutually exclusive and the term \textit{pornographic art} is an oxymoron \cite{levinson2005,uidhir2009}, while others consider the existence of grey areas between the two concepts \cite{patridge,vasilaki2010}. The assumptions at the base of our research consider that, while exceptions exist, there are classical dichotomies to distinguish \textit{prototypical cases} of artistic nudity vs pornography \cite{maes2011}, which include: \textit{subjectivity versus objectification}; the \textit{beautiful versus the smutty};
\textit{contemplation versus arousal}; the \textit{complex versus the one-dimensional}; the \textit{original versus the formulaic}; and \textit{imagination versus fantasy}. 
Focusing on some of the elements that characterize prototypical instances of pornography (\textit{e.g.}, being objectifying and formulaic), when compared to those characterizing artistic nudity (\textit{e.g.}, being subjective and original), our experimental design assumes that artistic nudity should not be censored on social media platforms and thus should be classified as \emph{safe} by NSFW classification algorithms. We also acknowledge that any criteria to differentiate between pornographic and artistic content constitutes an over-simplification and a discussion about the appropriateness of content moderation policies when applied to pornography is outside the scope of our research.  
Our focus is, instead, on analyzing the performance of machine learning models on artistic nudity with the purpose of mitigating existing limitations and thus contributing to the preservation of artistic freedom online. 

Thus, the main contributions of our work are four-fold:  
(1) We investigate the performance of three pre-trained NSFW classifiers on artistic nudity;
(2) We explore fine-tuning as a technique to improve the performance of the studied NSFW classifiers on artistic nudity;
(3) We illustrate the potential of considering multiple modalities to successfully address this challenge by means of a proof-of-concept with a multi-modal deep learning-based model (CLIP); 
and (4) We 
provide a reflection on this ethically complex and culturally relevant phenomenon.

\vspace*{-.1in}
\section{Models and Data}
In our experiments, we study the performance of three NSFW classifiers on three different datasets, described next.

\paragraph{\textbf{1. NSFW classifiers}}
Algorithms and models powering social platforms are proprietary and integrated into workflows involving humans. Hence, independent studies like ours are currently forced to use publicly available models as a proxy. While not ideal, this approximation is justified given that the technology behind these commercial models is believed to be similar, as reported in \cite{Dubettier2023}. 
Below, we summarize the characteristics of the three recent and openly accessible binary NSFW classifiers ("safe'' \textit{vs} "unsafe'' content) used in our experiments.

\textbf{- NudeNet}\footnote{Github Repository: \url{https://github.com/notAI-tech/NudeNet}, Last Access: 06.09.2023.} (\textsc{C01}) \cite{bhatti2018} consists of a \textsc{Resnet50} \cite{resnet50} convolutional neural network, pre-trained on 160,000 auto-labeled images (YahooNSFW classification model) and fine-tuned with their proprietary dataset.
When tested on their dataset with 2,000 images, 
the authors report 94.7\% accuracy. 

\textbf{- OpenNSFW2}\footnote{Github Repository: \url{https://github.com/bhky/opennsfw2}, Last Access: 06.09.2023.} (\textsc{C02}), consisting of a pre-trained deep neural network (\textsc{Resnet50}) on the ImageNet 1000-class dataset \cite{imagenet1000} and fine-tuned on a proprietary dataset of NSFW images. This is the model used by Yahoo!

\textbf{- Private Detector}\footnote{Github Repository: \url{https://github.com/bumble-tech/private-detector}, Last Access: 07.09.2023.} (\textsc{C03}), composed of a deep neural network pre-trained on proprietary, private data collected by the dating app Bumble \cite{bumble}. The model is based on the \textsc{Efficient Net V2} architecture \cite{efficientnetv2}.

\paragraph{\textbf{2. Datasets}}
We study the performance of the above models on  three datasets. 

\textbf{- D01: Censored Art Dataset.}
Given the proprietary nature of social media platforms, 
it is difficult to access datasets of censored art images. In fact, we are not aware of any publicly available dataset for this purpose. By means of a collaboration with \texttt{Don't Delete Art}, we were granted access to a diverse dataset of 143 images of contemporary art that (1) depict nudity and (2) had been censored on social media. \texttt{Don't Delete Art} is a group composed of NCAC's Arts \& Culture Advocacy Program\footnote{NCAC's Arts \& Culture Advocacy Program, \url{https://ncac.org/project/arts-culture-advocacy-program}, Last Access: 03.09.2024}, Artists at Risk Connection\footnote{Artists at Risk, \url{https://artistsatriskconnection.org/}, Last Access: 03.09.2024}, and Freemuse\footnote{Freemuse, \url{https://freemuse.org/}, Last Access: 03.09.2024}, along with artist-activists Emma Shapiro and Spencer Tunick, dedicated to protecting artistic expression online and to raising public awareness to the damage caused by social media companies censoring art. While the size of this dataset might seem limited, it is very difficult to gather larger datasets about this phenomenon. 
Despite its size, the data in \textsc{D01} is diverse from different perspectives: it contains images from almost 80 distinct artists, covering a 7-year period and spanning different artistic styles,  with 67\% of the images being either photographs or photorealistic drawings. Thus, we consider this dataset to be representative of the phenomenon under study. 

\begin{table}[t]
    \centering
    \caption{\textbf{Left:} Platforms and years where the images in dataset \textsc{D01} were censored. Note that several images were censored on different platforms and/or in different years. \textbf{Right:} Distribution of artworks in dataset \textsc{D02}. Blue bars: Distribution according to the gender of the depicted subjects in the artwork. Orange bars: Distribution according to the time period when the artwork was published.}
    \vspace{0.3cm} 
    \begin{tabular}{cc}
        \raisebox{-0.1\height}{ 
            \begin{minipage}[t]{0.48\textwidth}
                \centering
                \begin{tabular*}{\linewidth}{@{\extracolsep{\fill}}lcc|cc}
                    \toprule
                    \textit{Platform:} & \multicolumn{1}{c}{\tiny{\# samples}} &  & \textit{Year:} & \multicolumn{1}{c}{\tiny{\# samples}} \\
                    \midrule
                    Instagram & 80 & & 2016 & 2 \\
                    Facebook & 22 & & 2017 & 4 \\
                    Google & 2 & & 2018 & 10 \\
                    YouTube & 2 & & 2019 & 18 \\
                    HostGator & 1 & & 2020 & 22 \\
                    Tumblr & 2 & & 2021 & 29 \\
                    Whatsapp & 1 & & 2022 & 31 \\
                    TikTok & 1 & & 2023 & 18 \\
                    \textit{Unknown} & 53 & & \textit{Unknown} & 32 \\
                    \bottomrule
                \end{tabular*}
            \end{minipage}
        } &
        \raisebox{-0.6\height}{ 
            \begin{minipage}[t]{0.48\textwidth}
                \centering
                \includegraphics[width=\linewidth]{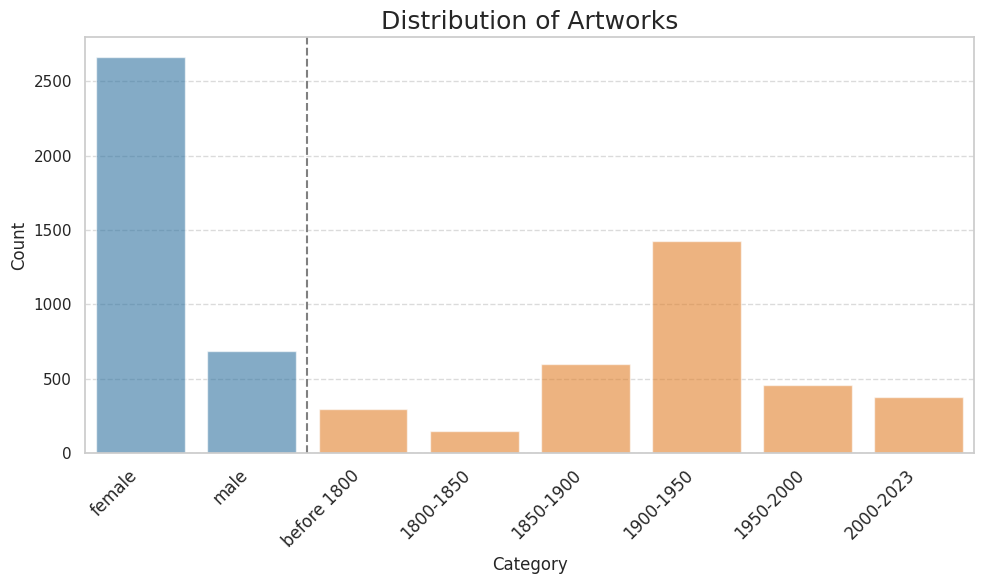}
            \end{minipage}
        }
    \end{tabular}
    \label{tab:dataset}
\end{table}

The images were censored over the span of seven years (from 2016 to 2023) and were provided to \texttt{Don't Delete Art} by the artists that created the images. Table \ref{tab:dataset} (left) summarizes the platforms and the years in which the images were censored. Instagram is the platform with the largest number of censored images, probably due to its popularity among artists. In addition, we observe an increasing number of available censored images in \textsc{D01} over time. This is probably due to a larger presence of artists on the platforms, the growing visibility of \texttt{Don't Delete Art} throughout the years, and the increasing reliance of the platforms on machine learning for content moderation. Figure \ref{fig:d01examples} depicts ten images that are part of this dataset.

\textbf{- D02: WikiArt Nudity Dataset.}
\textsc{D02} consists of 3,173 images from the WikiArt Online Collection\footnote{WikiArt, Last Access 29.12.23, \url{https://www.wikiart.org/}}, filtered according to the tags "male-nude" and 
"female-nude". The distribution of the images ---per gender and per time period--- is depicted in Table \ref{tab:dataset} (right).
There are \textbf{4x} more images representing female than male nudity, and the most represented historical period is the one spanning from 1900 to 1950, with almost 1,500 examples.

\begin{table}[t]
    \centering
    \caption{\textbf{Left:} Percentage of images classified as \textit{unsafe} by each of the three algorithms on the three analyzed datasets. The worst results are highlighted in red bold font. \textbf{Right:} Recall of the three classifiers on the three considered test sets before any fine-tuning process. The ground truth is as follows: all the images in \textsc{D01} and \textsc{T02} are labeled as "safe" and all the images in \textsc{T03} as "unsafe". Thus, in the case of \textsc{D01} and \textsc{T02}, the values correspond to the percentage of images that are classified as \textit{safe} whereas in the case of \textsc{T03} the values reflect the percentage of images that are considered to be \emph{unsafe}. Best result marked in green bold font.}
    \begin{tabular}{cc}
        \begin{subtable}[t]{0.48\textwidth}
            \centering
            \begin{tabular*}{\linewidth}{@{\extracolsep{\fill}}lccc}
                \toprule
                \textit{Case Study} & \multicolumn{1}{c}{\textsc{C01}} & \multicolumn{1}{c}{\textsc{C02}} & \multicolumn{1}{c}{\textsc{C03}} \\
                \midrule
                \textsc{D01 $\downarrow$} & 34.7\% & \textbf{\worst{47.9\%}} & 21.5\% \\
                \textsc{D02 $\downarrow$} & 8.0\% & \textbf{\worst{35.8\%}} & 7.4\% \\
                \textsc{D03 $\uparrow$} & 95.8\% & 94.7\% & \textbf{\worst{72.2\%}} \\
                \bottomrule
            \end{tabular*}
        \end{subtable} &
        \begin{subtable}[t]{0.48\textwidth}
            \centering
            \begin{tabular*}{\linewidth}{@{\extracolsep{\fill}}lccc}
                \toprule
                \textit{Case Study} Table 2: Left:& \multicolumn{1}{c}{\textsc{C01}} & \multicolumn{1}{c}{\textsc{C02}} & \multicolumn{1}{c}{\textsc{C03}} \\
                \midrule
                \textsc{D01 $\uparrow$} & 65.3\% & 52.1\% & \textbf{\best{78.5\%}} \\
                \textsc{T02 $\uparrow$} & \textbf{\best{91.7\%}} & 59.3\% & 89.6\% \\
                \textsc{T03 $\uparrow$} & \textbf{\best{95.2\%}} & 93.8\% & 74.5\% \\
                \bottomrule
            \end{tabular*}
        \end{subtable}
    \end{tabular}
    \label{tab:performance}
\end{table}

\textbf{- D03: NSFW Nudity Dataset.}
\textsc{D03} consists of 3,043 pornographic images from Reddit\footnote{Reddit, \url{https://www.reddit.com/}, Last Access: 19.01.2024}, obtained from 15 sub-reddits that explicitly contain professional and amateur pornography, without further details about the considered porn category. These images are intentionally recent (posted between the 24th of October 2022 and the 8th of November 2023) 
to minimize the probability that they were part of the training sets of any of the considered NSFW classifiers. 

\section{Content moderation on artistic nudity}

The evaluation experiments described in this section concern the three image datasets $D_i$ and the three  NSFW classifiers $f^{i}_\theta: D \to \mathbb{R}^d$ that map the input images to a $d$-dimensional output vector containing the assessment of the models regarding the NSFW nature of each image. 
In our case, $d = 1$ (binary classifiers).
The percentage of images classified as unsafe by each NSFW classifier on each dataset is summarized in Table \ref{tab:performance} (left). 
All the images in the Censored Art (\textsc{D01}) and the WikiArt Nudity datasets (\textsc{D02}) correspond to artworks contributed by artists. As previously explained, we consider all artistic depictions of nudity to be safe. As a consequence, all the images that are labeled as unsafe in these datasets are considered to be false positives. Depending on the model, the false positive rate ranges from 21.5\% to 47.9\% on \textsc{D01}, and from 7.44\% to 35.8\% on \textsc{D02}. In both cases (\textsc{D01} and \textsc{D02}), the NSFW classifiers that yield the largest / smallest number of false positives are \textsc{C02} and  \textsc{C03}, respectively. However, we observe that \textsc{C03} only considers unsafe 72.16\% of the images in \textsc{D03}. Thus, we conclude that this model censors fewer artworks not because of a better ability to distinguish pornographic \textit{vs} artistic nudity but because it is generally more permissive towards nudity. Interestingly, the analyzed classifiers have significantly larger false positive rates on the images in \textsc{D01}  (contemporary censored art) when compared to the images in \textsc{D02} (WikiArt) (Mann-Whitney U Statistic test, \textsc{C01}, p<0.01; 
\textsc{C02}, p<0.01; 
and\textsc{C03}, p<0.001). 

While \emph{all} the images in \textsc{D01} had been already censored on social media, only a portion of them is also censored by the models considered in this study. This might be due to an improvement of the NSFW algorithms throughout the years, hence becoming more \textit{art-aware}. However, it might also hint that social media platforms use more conservative models with higher false positive rates and/or apply specific policies regarding artistic nudity according to internal governance, economic and/or ideological reasons. Interestingly, the three NSFW classifiers also exhibit significantly different performances on the images of \textsc{D01}. 
While being based on similar deep learning architectures, these models were trained on \textit{different datasets}, leading to different learned representations, particularly if a different ground truth labeling system was used in the training process. 

In the next section, we delve deeper into the performance of the NSFW classifiers to shed light on their potential biases.

\subsection{Analysis of Biases}

\begin{table}[ht]
\caption{Percentage of false positives (images classified as \textit{unsafe}) per gender and time period by each of the three algorithms on the WikiArt Nudity dataset (\textsc{D02}). The per-gender worst results are highlighted in red bold font.}
\centering
\begin{tabular*}{\columnwidth}{@{\extracolsep{\fill}}lccc}
\toprule
\textit{WikiArt dataset} & \multicolumn{1}{c}{\textsc{C01}} & \multicolumn{1}{c}{\textsc{C02}} & \multicolumn{1}{c}{\textsc{C03}} \\
\midrule
Overall & 8.0\% 
& 35.8\% & 7.4\% \\ 
\quad \quad Female & \textbf{\worst{8.3}}\% & 35.0\% & \textbf{\worst{7.7\%}} \\
\quad \quad Male & 5.5\% & \textbf{\worst{35.7}}\% & 4.5\% \\

\bottomrule
Female - Male (\%) & & & \\
\quad \quad before 1800 & \small \textbf{\worst{10.3}} - 8.0 & \small \textbf{\worst{50.8}} - 42.0 & \small \small \textbf{\worst{9.7}} - 7.3  \\
\quad \quad 1800-1850 & \small \textbf{\worst{13.4}} - 9.8 & \small 45.5 - \textbf{\worst{55.7}} & \small \textbf{\worst{6.2}} - 3.3 \\
\quad \quad 1850-1900 & \small \textbf{\worst{11.8}} - 8.4 & \small 38.9 - \textbf{\worst{44.3}} & \small \textbf{\worst{9.7}} - 6.1  \\
\quad \quad 1900-1950 & \small \textbf{\worst{7.8}} - 3.5 & \small \textbf{\worst{36.9}} - 34.0 & \small \textbf{\worst{8.1}} - 2.5 \\
\quad \quad 1950-2000 & \small \textbf{\worst{4.9}} - 4.2 & \small 29.8 - \textbf{\worst{31.0}} & \small 5.6 - 5.6 \\
\quad \quad 2000-2023 & \small \textbf{\worst{7.9}} - 1.0 & \small \textbf{\worst{20.6}} - 17.9 & \small \textbf{\worst{5.6}} - 2.1
\\
\bottomrule
\label{tab:percgender}
\end{tabular*}
\end{table}

\paragraph{Sensitivity to gender and time period}
Table \ref{tab:percgender} reports the percentage of false positives of each of the models on the WikiArt dataset (\textsc{D02}), depending on the gender and time period of the artwork. Regarding the time period, the largest false positive rates correspond to images prior to the 20th century. Regarding gender, the false positive rates of \textsc{C01} and \textsc{C03} are significantly larger for images depicting females than males (Mann-Whitney U Statistic test, p<0.01 and p<0.05, respectively).

\begin{figure*}[t]
\centering
\begin{subfigure}{\linewidth}
\centering 
\includegraphics[width=\linewidth]{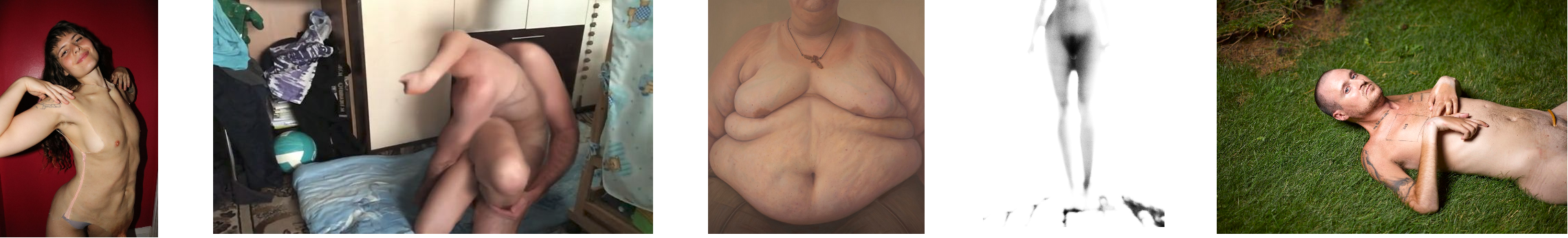}
\caption{\textit{Authors of the images (from left to right): Manuela Benaim, Santina Amato, Clarity Haynes, Heather M of the Femme Project, Robert Andy Coombs.}}
\end{subfigure}

\begin{subfigure} {\linewidth}
\centering
\includegraphics[width=\linewidth]{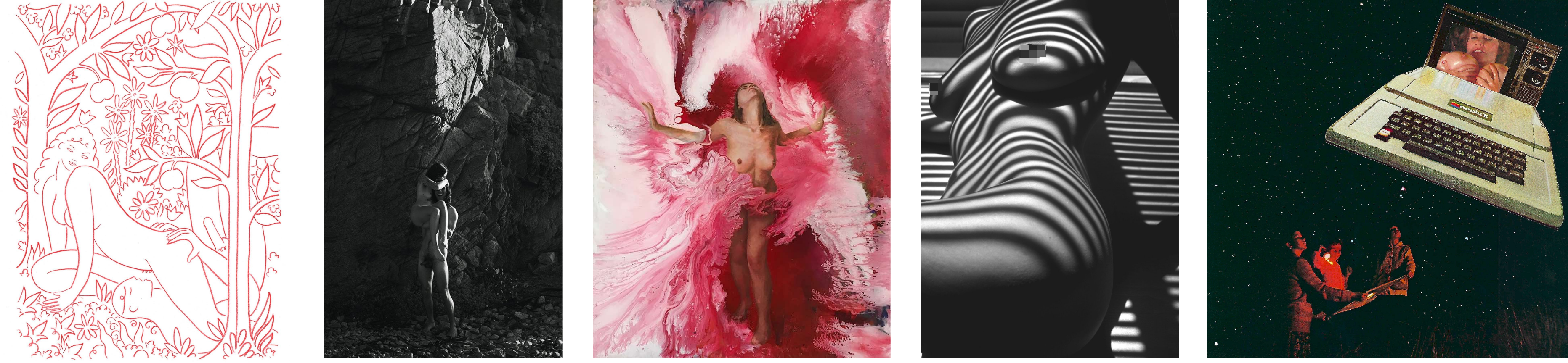}
\caption{\textit{Authors of the images (from left to right): Alphachanneling, Danilo Garrido, Annata Bartos, Savannah Spirit, Justin Eldridge.}} 
\end{subfigure}
  
  \caption{Exemplary images in \textsc{D01} that are considered to be \textit{unsafe} (top) or \textit{safe} (bottom) by the three NSFW classifiers.}
  \label{fig:d01examples}
\end{figure*}

\begin{figure*}[t]
\centering
\begin{subfigure}{\linewidth}
\centering 
\includegraphics[width=\linewidth]{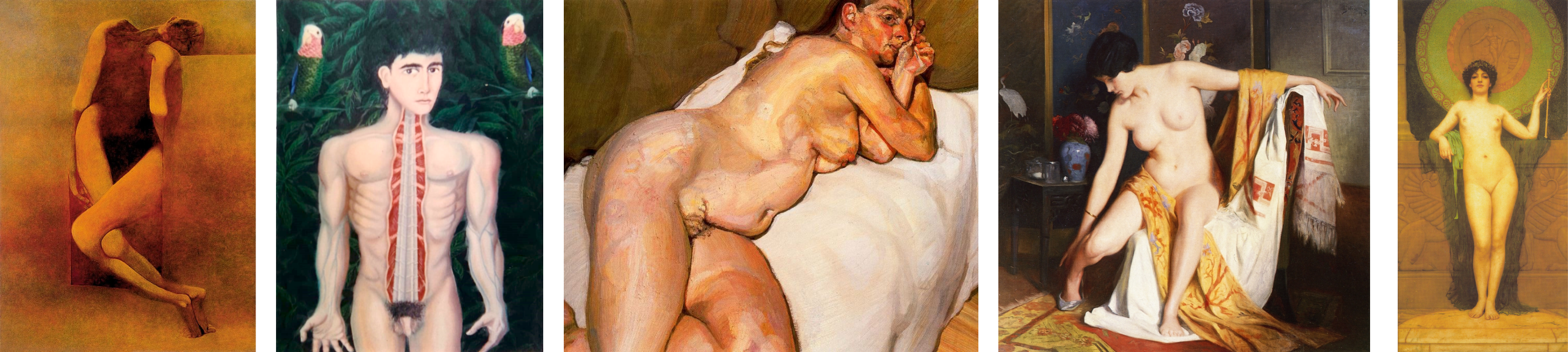}
\caption{\textit{From left to right: \textit{Untitled} (Zdzislaw Beksinski), \textit{Anatomic Study with Parrots} (Enrique Silvestre), \textit{Naked woman on a sofa} (Lucian Freud), \textit{Nude in an interior} (Julius LeBlanc Stewart), \textit{Campaspe} (John William Godward).}}
\end{subfigure}

\begin{subfigure} {\linewidth}
\centering
\includegraphics[width=\linewidth]{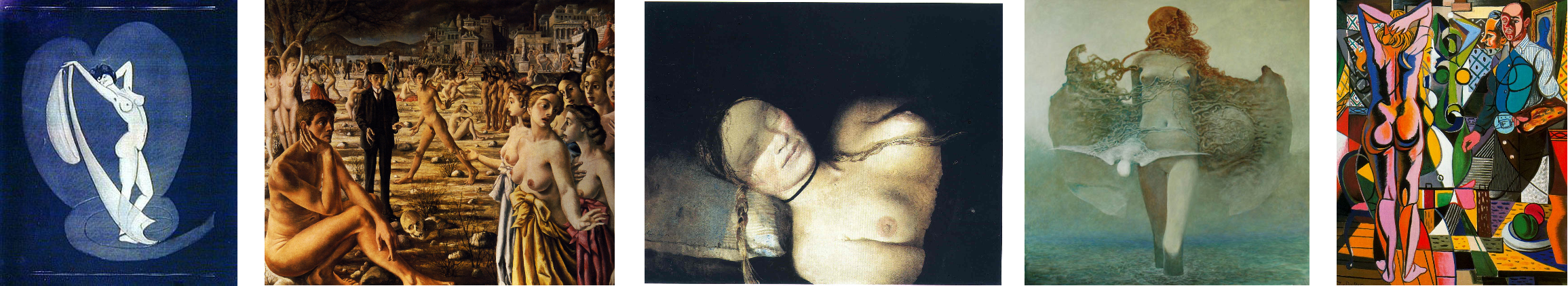}
\caption{\textit{From left to right: \textit{Salome} (John Vassos), \textit{City worried} (Paul Delvaux), \textit{Untitled} (Andrew Wyeth), \textit{Untitled} (Zdzislaw Beksinski), \textit{Self-portrait with model and the still life} (Rafael Zabaleta).}} 
\end{subfigure}
  
  \caption{Exemplary images in \textsc{D02} that are considered to be \textit{unsafe} (first line) or \textit{safe} (second line) by all the three models.}
  \label{fig:d02examples}
\end{figure*}

\paragraph{Inter-algorithm analysis}
The behavior of the three classification algorithms is not consistent when tested on the same dataset, yielding different false positive rates. 
We identified the images from the art-related datasets (\textsc{D01} and \textsc{D02}) on which there was agreement on the decisions by \textit{all} the models. In \textsc{D01}, 5 images were considered to be \textit{unsafe} by all the models and 55 images were considered to be \textit{safe}. Examples for both sets of images are provided in Figure \ref{fig:d01examples}. The two sets of images do not differ in terms of semantic "explicitness", but the censored images tend to depict human bodies in a rather central position, surrounded by fewer artifacts and artistic elements than the uncensored ones. In the case of \textsc{D02}, a total of 81 images were considered to be \textit{unsafe} by the three models and 1,921 were considered to be \textit{safe} (examples are reported in Figure \ref{fig:d02examples}). Among the 81 artworks that were considered to be unsafe, 75 display at least one female body (92.6\%) and 11 display at least one male body (13.6\%). Considering the time period, 44 images (54.3\%) belong to 1900-1950, 18 images (22.2\%) belong to 1850-1900, 8 images (9.88\%) belong to before 1800 and 2000-2023, finally 2 images (2.47\%) to 1950-2000 and 1 image (1.23\%) to 1800-1850. These percentages approximately correspond to the proportions depicted in Table \ref{tab:dataset} (right), which represent the corresponding rates for the whole dataset. 

\paragraph{Sensitivity to artistic style} According to previous qualitative work \cite{riccio2024}, certain artistic styles seem to be more likely to be censored than others. Hinted by this finding, we performed a per-artist analysis of the 81 images in \textsc{D02} that were labeled as \textit{unsafe} by the three NSFW classifiers. Such images belong to 50 distinct, unique authors. The most censored artist is Zinaida Serebriakova, with 11 (13.6\% of the 81 total images) of her artworks classified as \textit{unsafe} by the three models. This is a disproportionate percentage given that only 53 of her paintings are part of the total dataset (less than the 2\%). The number of artworks by other authors with a similar presence in the dataset that are classified as \textit{unsafe} is significantly smaller than in the case of Serebriakova: for instance, there are 54 artworks by Amedeo Modigliani in \textsc{D02}, but only 3 of them are classified as \textit{unsafe} by all the models. These findings empirically corroborate the hypothesis that certain artistic styles are more likely to be censored than others. 

Given the limitations of the NSFW classifiers when it comes to discerning between artistic and pornographic nudity, we explore next the capabilities of fine-tuning as a suitable approach to make these models more \textit{art-aware}. 

\subsection{Fine-tuning}

Fine-tuning has been found to be a powerful approach to enhance the performance of pre-trained machine learning models, also in the case of fine art classification \cite{cetinic2018}. The process of fine-tuning leverages the knowledge acquired by a model when trained on a large, diverse and generic dataset. By focusing on a more specific domain or problem, fine-tuning allows the pre-trained models to adapt the learned features and representations to the nuances of the target task. Fine-tuning is particularly valuable and effective when there is limited labeled data for the target task ---as in our case--- because it enables transferring the general knowledge of the pre-trained models to the new task. The three classifiers are pre-trained models that we fine-tune with a small dataset corresponding to the task at hand, \textit{i.e.}, the correct classification of pornographic \textit{vs} artistic nudes. We describe next the details of our fine-tuning process and the obtained results.

\paragraph{Implementation}
We considered all the images (N=143) in \textsc{D01} as a test set. Furthermore,  we randomly sampled 145 images (to roughly match the size of \textsc{D01}) from \textsc{D02} and \textsc{D03} to create two additional test sets (\textsc{T02} and \textsc{T03}). 
The remaining images in \textsc{D02} and \textsc{D03} were used as training and validation sets of the fine-tuning process. 
The training sets were divided into 5 different folds containing 20\% of the images. In each experiment we selected four folds (80\% of both sets) as training and one fold (20\% of both sets) as validation, and performed the experiments five times. For the fine-tuning process, we followed the guidelines available on the Github repositories where each of the models were available. In the case of \textsc{C01} and \textsc{C02}, all the layers of the model but the last one were frozen such that only the last layer was fine-tuned\footnote{More details available at: "Transfer Learning \& Fine Tuning", Keras, \url{https://keras.io/guides/transfer_learning/}, Last Access: 08.02.2024.}. In the case of \textsc{C03}, and according to the guidelines, we simply continued training the model with the fine-tuning training data. 

\begin{figure*}[t]
    \centering
    \begin{subfigure}[t]{0.48\textwidth}
        \centering \vspace{-1.3in}\includegraphics[width=\linewidth]{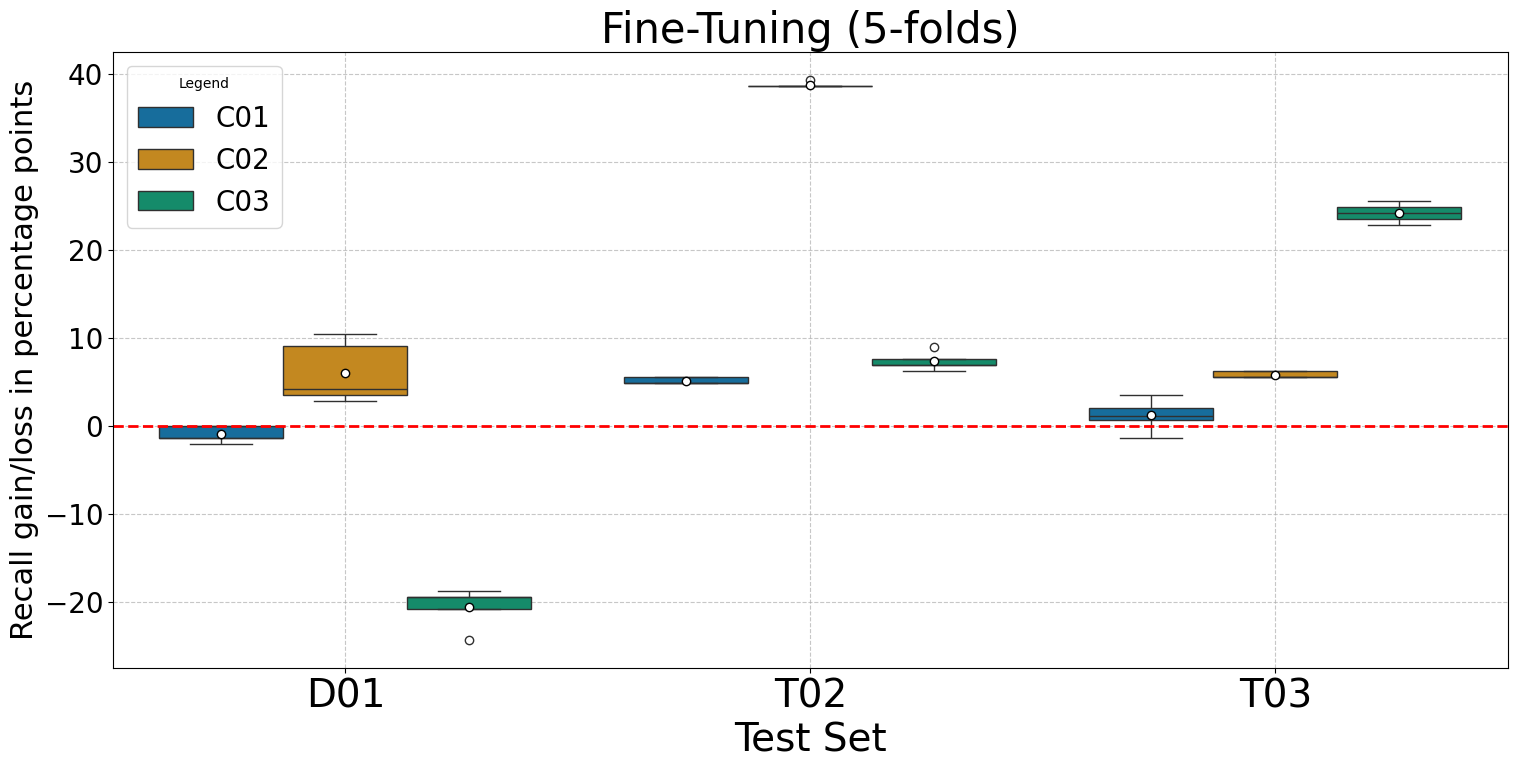}
    \end{subfigure}
    \hfill
    \begin{subfigure}[t]{0.48\textwidth}
        \centering
        \includegraphics[width=\linewidth]{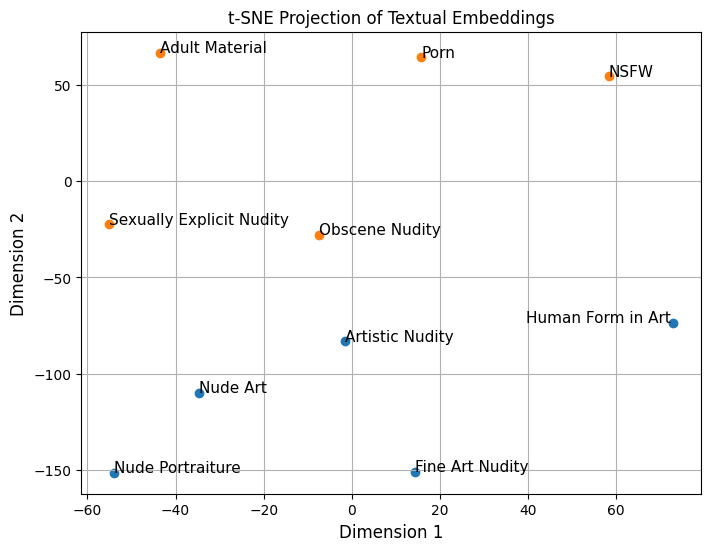}
    \end{subfigure}
    \caption{\textbf{Left:} Recall gain/loss (in percentage points) on each of the three test sets after fine-tuning each of the three NSFW classifiers. The results are shown as boxplots with the mean (white dot) and the standard deviation (bars) of the recall gain/loss over the 5 considered folds. \textbf{Right:} t-SNE projection of the CLIP textual embeddings of the considered terms in $S_{porn}$ and $S_{art}$ with PCA initialization. The existence of two clusters is confirmed via k-means.}
    \label{fine}
\end{figure*}

\paragraph{Results} 
The initial performance of the three models on the three test sets is reported in Table \ref{tab:performance} (right), where we provide the recall of the algorithms on each dataset ---\textit{i.e.,} the percentage of images in \textsc{D01} and \textsc{T02} that are classified as \emph{safe}, and the percentage of the images in \textsc{T03} that are considered to be \emph{unsafe}---. The effect of the fine-tuning is summarized in Figure \ref{fine} (left), depicting the mean and standard deviation of the performance gain/loss (in percentage points) for each of the fine-tuned classifiers on each of the test sets.

After fine-tuning, we observe an improvement in the performance of the three NSFW classification algorithms on \textsc{T02} and \textsc{T03}, stabilizing at above 95\%. However, on \textsc{D01}, the behavior of the three models differs significantly. 
In the case of \textsc{C01}, the recall value shifts from 65.3\% to an average of 64.3\%, with a decrease of 1 percentage point; in the case of \textsc{C02}, the recall value shifts from 52.1\% to an average of 57.9\%, with an improvement of 5.7 percentage points; and in the case of \textsc{C03}, the recall value shifts from 78.5\% to an average of 57.9\%, with a decrease of 20.6 percentage points.
As a result, the percentage of images from \textsc{D01} that are classified as \emph{safe} stabilizes around 60\% for the three analyzed classifiers. 
Given these limitations in performance and the lack of consistency among the three NSFW classifiers, we conclude that visual information might not be sufficient to correctly discern the artistic nudity in \textsc{D01} from pornography.

\section{Zero-Shot Multi-modal Classification}
In this section, we explore the potential of combining two modalities (images and text) to address the limitations of image-based NSFW classifiers regarding their ability to correctly discern between artistic and pornographic nudity, even after fine-tuning.  Multi-modal systems have been found to facilitate contextual reasoning \cite{awais2023}, and recent research has highlighted the need of considering contextual information to correctly distinguish between artistic and pornographic nudity \cite{riccio2022}. We consider the Contrastive Language-Image Pre-training model or CLIP \cite{radford2021}. CLIP is part of a family of deep learning models that leverage contrastive learning \cite{chen2020}, a training method where the model learns to distinguish between positive (correct associations) and negative (incorrect associations) pairs by incorporating modality-specific encoders for both images and text, and generating embeddings for each modality in the same latent representation. During training, a contrastive loss is employed to enhance the alignment between the embeddings for pairs of images and text, allowing it to generalize well across various applications, such as image classification, object detection, and zero-shot classification. In zero-shot classification, a model is employed to recognize classes that have never been seen during training. This is achieved by leveraging auxiliary information about the classes, allowing the model to predict the class of unseen examples based on similarities to the auxiliary information \cite{norouzi2013,xian2018}. In the case of zero-shot image classification through CLIP, the auxiliary information is provided in the form of textual descriptions at inference time. The classification process is based on finding matches between the provided description and the images, as described next.


\paragraph{Implementation}
Given the three image datasets $D_i$ and two sets of textual terms, $S_{porn}$ and $S_{art}$, describing pornography and artistic nudity respectively, we use a pre-trained CLIP to perform zero-shot classification of the images in $D_i$. 
CLIP is a combination of two encoders $f_\theta: D \to \mathbb{R}^d$ and $f_\gamma: S \to \mathbb{R}^d$ that map input images in $D$ and input texts in $S$ to the same latent space of dimension $d$. Given an image from $D$, its classification as \emph{safe} or \emph{unsafe} is performed according to the Algorithm in Table \ref{tab:zeroshot} (left), \emph{i.e.}, it is based on the distance of the image embedding  to the text embeddings. As reflected in the Algorithm, different combinations of the terms in $S$ are considered yielding a set of accuracies from which the mean accuracy and its standard deviation are computed. The $kNN$ algorithm corresponds to the weighted kNN provided by the \textsc{scikitlearn} Python library, with k equal to the number of available text embeddings in the considered combination of textual terms ($S_i$), and using cosine similarity as the weighting metric. We use the backbone architecture \textsc{convnext\_base\_w} pre-trained on \textsc{laion2b\_s13b\_b82k\_augreg} (default settings according to the open-source Github Repository OpenCLIP\footnote{OpenCLIP, \url{https://github.com/mlfoundations/open_clip}, Last Access: 05.02.2024}), with $d=640$.

In our experiments, $n=5$, $S_{porn}$ = \textit{"Porn, Sexually Explicit Nudity, Obscene Nudity, Adult Material, NSFW"} and $S_{art}$ = \textit{"Artistic Nudity, Nude Art, Fine Art Nudity, Nude Portraiture, Human Form in Art"}. These textual terms were chosen based on our domain knowledge of the field. As illustrated in Figure \ref{fine} (right), they are separable in CLIP's latent space after t-SNE projection. 
The combinations of textual embeddings that compose $S$ in the Algorithm in Table \ref{tab:zeroshot} (left) include the same number of textual terms from $S_{porn}$ and $S_{art}$. For example, two possible textual combinations are \{ \textit{"Fine Art Nudity", "Porn"} \} and \{ \textit{"Artistic Nudity", "Nude Portraiture", "Porn", "Obscene Nudity"} \}.

\begin{table}[ht]
    \centering
    \caption{\textbf{Left:} Zero-Shot Multi-Modal classification algorithm \textbf{Right:} Recall of the multi-modal approach on the three datasets with respect to $k$. The value of $k$ represents the number of textual embeddings in the considered combination and the number of neighbors in the kNN.}
    \vspace{0.3cm} 
    
    \begin{tabular}{cc}
        \begin{minipage}[t]{0.48\textwidth}
            \centering
            \vspace{-0.2cm} 
            \begin{algorithmic}[1]
                \REQUIRE  \quad \\
                Dataset $D$  (\textsc{D01}, \textsc{D02}, \textsc{D03}), ground truth $y$, set of textual terms $S_{porn}$, $S_{art}$, $n=|S_{porn}|=|S_{art}|$, encoders $f_\theta: D \to \mathbb{R}^d$ and $f_\gamma: S \to \mathbb{R}^d$ 
                \ENSURE \quad \\
                $S = \bigcup\limits_{i=1}^{n} \{ P \cup A \; | \; P \subset S_{porn}, $ \\ $ \quad \quad \quad \quad A \subset S_{art}, \; |P|=|A|=i \}$\\
                \STATE $ e_{D} \leftarrow f_\theta(D) $ \\
                \STATE $ A \leftarrow \{ \}$
                \FOR{$i=1$ \textbf{to} $ |S|$} 
                \STATE $ e_{S_i} \leftarrow f_\gamma(S_i) $
                \STATE $ \hat{y} \leftarrow kNN (e_{S_i}, e_{D}) $ with $k = |S_i|$
                \STATE $ A \leftarrow A\cup\{acc(y, \hat{y})\}$
                \ENDFOR
                \RETURN A
            \end{algorithmic}
        \end{minipage}
        &
        \begin{minipage}[t]{0.48\textwidth}
            \centering
            \vspace{-0.1cm} 
            \includegraphics[width=\linewidth]{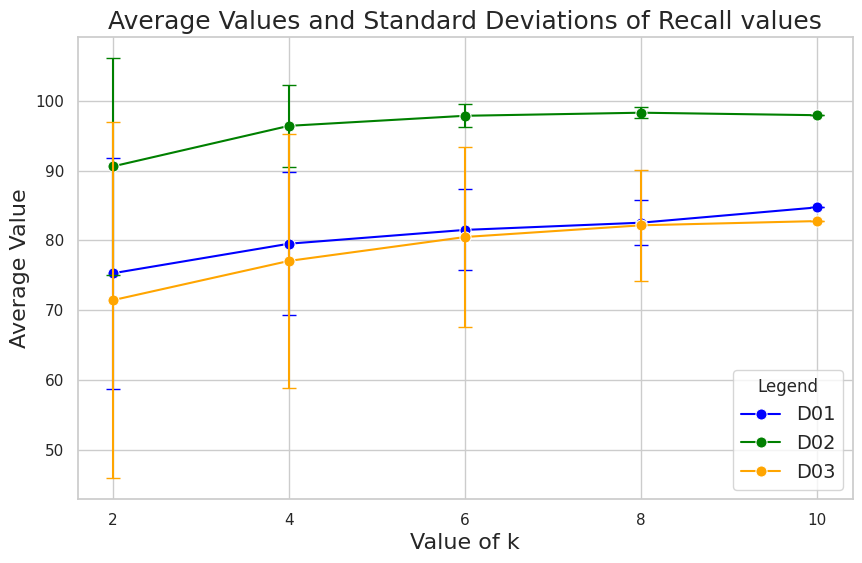}
        \end{minipage}
    \end{tabular}
    \label{tab:zeroshot}
\end{table}

\paragraph{Results}
Table \ref{tab:zeroshot} (right) depicts the mean/std recall values on the three datasets obtained by means of the Algorithm in Table \ref{tab:zeroshot} (left) with the previously explained textual terms, $S_{porn}$ and $S_{art}$. Note how the performance improves with \textit{k} which is the number of textual embeddings in the considered textual combination $S_i$, reaching \textbf{84.7\%} on D01, \textbf{97.9\%} on D02 and \textbf{82.8\%} on D03 when $k=10$. Comparing these results with those reported in Table \ref{tab:performance}, we observe a significant improvement on the artistic data, particularly on \textsc{D01}, the dataset of censored of contemporary artists. In this case, the performance is \textbf{29.7\%, 62.6\%} and \textbf{8\%} better than the original performance of \textsc{C01}, \textsc{C02} and \textsc{C03}, respectively. The performance achieved on \textsc{D02} is also remarkable, representing an improvement of $6.8\%$, $65.1\%$ and $9.3\%$ when compared to the original performance of \textsc{C01}, \textsc{C02} and \textsc{C03}, respectively. 
Finally, regarding \textsc{D03}, a recall of 82.8\% represents an improvement of $14.7\%$ of \textsc{C03}'s original performance, yet it is lower than that the performance of \textsc{C01} and \textsc{C02} on this dataset.  Interestingly, a visual inspection of the misclassified images in \textsc{D03} reveals that none of them depicts sexual intercourse and mostly contain female models in rather refined poses and lighting atmospheres. In this proof-of-concept, we find that multi-modal learning outperforms fine-tuned uni-modal approaches on this task, consistent with recent theoretical work on this topic \cite{lu2023}.

\vspace*{-.13in}
\section{Discussion}
From our analyses, we draw several implications that we hope will inform future research on the automatic moderation of artistic nudity.  

With false positive rates ranging between $21.5\%$ and $47.9\%$, the considered NSFW classifiers are unable to correctly discern between artistic and pornographic nudes. This poor performance might translate into artworks being censored online, with severe economic, professional and personal consequences for their creators \cite{riccio2024}.  
Investigating the algorithmic censorship of artistic nudity on social media involves considering a complex phenomenon shaped by the power of today's social media platforms \cite{baym2018,petre2019,cotter2023,hill2019}. The treatment of artistic nudity as pornography 
also raises questions about the cultural influence of the technology giants \cite{Poell2019,McCabe2024}. 
With a prominent role in today's art world \cite{riccio2022}, social media platforms determine which art is acceptable, which results on the censorship of artistic pieces without considering the historical and cultural significance of nudity in art as a form of expression \cite{lynda2022}.

Artistic expression is not solely represented by the final product, as it also consists of the process of translating emotions and abstract ideas, or life experiences into tangible forms \cite{blumenfeld2016}. However, 
when machine learning models are used to moderate artistic content, they reduce it to a mere visual output regardless of its intrinsic creative depth, objectifying the meaning of art. 
Furthermore, the behavior of the tested NSFW classification algorithms is inconsistent when evaluated on the same datasets, yielding different false positive rates and being sensitive to gender and style. 
Thus, we conclude that the visual information alone does not seem to be sufficient to correctly perform this classification task, as illustrated by the results of our fine-tuning experiments. Indeed, our work emphasizes the lack of \textit{contextualization} and excessive \textit{literalization} \cite{leung2022} as one of the main pitfalls in contemporary content moderation practices.  

While this limitation is difficult to overcome with a strictly technical solution, multi-modal models, such as CLIP, show promise as a more flexible and context-rich approach to tackle this challenge. Considering that the difference between artistic and pornographic nudity is, in some cases, debatable \cite{vasilaki2010}, an interesting future research direction entails analyzing how humans perform in classifying the images in our datasets as artistic \textit{vs} the pornographic nudity, creating a "human"  benchmark for this nuanced task. 
In this direction, the proposed multi-modal approach allows for the inclusion of expert knowledge into the NSFW classification process, with the possibility of consulting with art experts to identify the relevant concepts and dimensions (auxiliary information) to consider when assessing the artistic value of an image (\textit{e.g.,} the pose, the lighting). 
CLIP, or similar multi-modal approaches, would enable the consideration of such dimensions, resulting in more explainable and human-centric NSFW classifiers.

For the authors' positionality and limitations of our research, we refer the reader to the Appendix.

\section{Conclusion}
In this paper, we have studied the algorithmic censorship of art on social media by analyzing the performance of three NSFW classifiers on artistic nudity. Our experimental results have revealed significant technical limitations in the algorithms' ability to discern between artistic and pornographic nudity based solely on visual information, even after fine-tuning. We have also identified the existence of a gender and a stylistic bias in the models' performance. To mitigate existing limitations on the classification of artistic nudity, we have proposed a novel multi-modal zero-shot classification approach.

\section*{Acknowledgments}
We are grateful to \texttt{Don’t Delete Art} for the fruitful collaboration and their willingness
to support our research. PR and NO are supported by a nominal grant received at the
ELLIS Unit Alicante Foundation from the Regional Government
of Valencia in Spain (Convenio Singular signed with Generalitat
Valenciana, Conselleria de Innovación, Industria, Comercio y Tur-
ismo, Dirección General de Innovación). PR is also supported by a grant by the Bank Sabadell Foundation. A part of this work was
performed while PR was an academic guest at ETH Zürich, in the
Data Analytics Lab. Her stay was partially supported by ELISE (GA
no 951847) and partially supported by ETH Zürich. GC acknowledges travel support from the European Union’s Horizon 2020 research and innovation programme under Grant Agreement No 951847.

%
%
\bibliographystyle{splncs04}
\bibliography{main}

\clearpage
\section*{Appendix}
\subsection*{Limitations} While providing interesting and unprecedented insights on the topic of algorithmic censorship of nudity, we reflect next about some of the limitations of our work. 

A first limitation is the size of the datasets used in our experiments, particularly \textsc{D01}. However, as previously noted, we are not aware of any publicly available dataset of censored art on social media. The dataset shared with us by \texttt{Don't Delete Art} is the largest dataset of this kind known to us. 
A second limitation of this study concerns access to our datasets. The dataset of censored art (\textsc{D01}) is not publicly available as we obtained access to it by means of our collaboration with \texttt{Don't Delete Art}. The WikiArt dataset (\textsc{D02}) is publicly available. The third dataset (\textsc{D03}) is not publicly available due to privacy.
The third limitation relates to the analysis of biases. We only focused on the image attributes that we could easily access (\textit{e.g.}, the presence of female \textit{vs} male bodies in the images of \textsc{D02}). However, there are other biases of interest that could be explored after manually labelling the images in the dataset. Future work could consider whether specific artistic media (\textit{e.g.,} photos \textit{vs} paintings) or artistic movements (\textit{e.g.,} impressionism \textit{vs} expressionism) are more likely to be censored than others. We empirically observed that the images in \textsc{D02} were significantly less likely to be considered \emph{unsafe} by the algorithms when compared to the images in \textsc{D01}, yet the reasons for this difference in performance remain unclear. It could be due to the specific aesthetics and artistic medium of the images in \textsc{D01}, or to the popularity of some of the images in \textsc{D02}, which might have been included in the training sets of the considered models. 

\subsection*{Authors' Positionality}
Given the subject matter of this study, it is important to highlight the authors' positionality and potential subjectivity in this research. At the time of the study, two of the authors are researchers in a research foundation devoted to the study of human-centric and responsible AI for Social Good, and the other two are scholars in reputed European and American universities. Three of the authors identify as female and one as male. We are originally of three different European nationalities and have spent many years working abroad in different cultural contexts. Our core research areas are Artificial Intelligence, Mobile and Ubiquitous Computing, Computational Social Sciences, Computational Creativity, Human-computer Interaction, AI Ethics and AI for Social Good. Our multidisciplinary background, including both technical and ethics expertise, enabled us to analyze the technical aspects and impact of algorithmic censorship of artistic nudity, while deeply understanding its ethical implications. However, none of the authors had firsthand experience with algorithmic censorship of artistic content, and all reside in open, democratic societies. For this reason, our partnership with \texttt{Don't Delete Art} proved to be essential to deepen our understanding and gather valuable feedback for our research.

\end{document}